%% file: main.tex
\documentclass[10pt,twocolumn]{article}
\input{macro}

\iccvfinalcopy

\def\iccvPaperID{1244} 

\ificcvfinal\fi

\begin{document}

\title{\textcolor{mycolor}{\paris}: \textcolor{mycolor}{Pa}rt-level \textcolor{mycolor}{R}econstruction and Motion Analys\textcolor{mycolor}{is} for Articulated Objects}
\author{
\textbf{Jiayi Liu, Ali Mahdavi-Amiri, Manolis Savva}\\
Simon Fraser University\\
\url{https://3dlg-hcvc.github.io/paris/}}
\maketitle
\ificcvfinal\thispagestyle{empty}\fi

\begin{abstract}
\input{sections/00_abstract.tex}
\end{abstract}

\input{sections/01_introduction.tex}
\input{sections/02_related_work.tex}
\input{sections/03_problem_statement.tex}

\input{sections/04_methods.tex}
\input{sections/05_experiments.tex}

\input{sections/06_conclusion.tex}

\mypara{Acknowledgements.}
This work was funded in part by a Canada Research Chair grant (CRC-2019-00298), NSERC Discovery grants (RGPIN-2019-06489, RGPIN-2022-03111), and NSERC Discovery Launch Supplement (DGECR-2022-00359), and enabled in part by support from \href{https://www.westgrid.ca/}{WestGrid} and \href{https://www.computecanada.ca/}{Compute Canada}.
We thank Hanxiao Jiang, Han-Hung Lee, Yongsen Mao, Yilin Liu for helpful discussions, as well as Angel X. Chang, Sanjay Haresh, Ning Wang, and Qirui Wu for proofreading.

{\small
\bibliographystyle{plainnat}
\setlength{\bibsep}{0pt}
\bibliography{main}
}

\end{document}


\title{\textcolor{mycolor}{\paris}: \textcolor{mycolor}{Pa}rt-level \textcolor{mycolor}{R}econstruction and Motion Analys\textcolor{mycolor}{is} for Articulated Objects\\Supplemental Materials}
\author{\textbf{Jiayi Liu, Ali Mahdavi-Amiri, Manolis Savva}\\
Simon Fraser University\\}
\maketitle
\ificcvfinal\thispagestyle{empty}\fi

\appendix
\input{supplement/supplemental}

{\small
\bibliographystyle{plainnat}
\setlength{\bibsep}{0pt}
\bibliography{main}
}

%% file: macro.tex
\usepackage{iccv}
\usepackage[numbers,compress]{natbib}
\usepackage{times}
\usepackage{epsfig}
\usepackage{graphicx}
\usepackage{amsmath}
\usepackage{amssymb}
\usepackage{booktabs}
\usepackage{multirow}
\usepackage{comment}
\usepackage{xcolor}
\usepackage{subfigure}
\usepackage{comment}
\usepackage[pagebackref=true,breaklinks=true,colorlinks,bookmarks=false]{hyperref}
\usepackage{cleveref}
\usepackage{graphicx}

\newcommand{\denselist}{\itemsep 0pt\parsep=0pt\partopsep 0pt}
\newcommand{\mypara}[1]{\noindent\textbf{#1}}

\newcommand{\paris}{PARIS\xspace}

\definecolor{mycolor}{RGB}{118, 157, 213}

%% file: sections/00_abstract.tex
We address the task of simultaneous part-level reconstruction and motion parameter estimation for articulated objects. Given two sets of multi-view images of an object in two static articulation states, we decouple the movable part from the static part and reconstruct shape and appearance while predicting the motion parameters. To tackle this problem, we present \emph{\paris}: a self-supervised, end-to-end architecture that learns part-level implicit shape and appearance models and optimizes motion parameters jointly without any 3D supervision, motion, or semantic annotation. Our experiments show that our method generalizes better across object categories, and outperforms baselines and prior work that are given 3D point clouds as input. Our approach improves reconstruction relative to state-of-the-art baselines with a Chamfer-L1 distance reduction of 
$3.94$ ($45.2\%$)
for objects and 
$26.79$ ($84.5\%$)
for parts, and achieves 5\% error rate for motion estimation across 10 object categories.

%% file: sections/01_introduction.tex
\section{Introduction}
\label{sec:introduction}

Articulated objects consist of interconnected static and movable parts that exhibit motion.
Such objects are ubiquitous in real life (e.g., drawers, ovens, chairs, laptops, staplers).
Thus, perception and understanding of articulated object structure is important in many areas including robotics~\cite{Xiang_2020_SAPIEN, mao2022multiscan, fan2023arctic, liu_akb48}, animation~\cite{sia_motionrepre, goes_characterarti}, and industrial design~\cite{Jain_2019}.
Articulated object motion analysis enables robots to manipulate objects more effectively~\cite{Gadre_2021_ICCV}.
Acquiring digital replicas of articulated objects~\cite{jiang_ditto:_2022, hsu_dittohouse} also enables simulating object articulation in applications involving robotic agents and embodied AI~\cite{savva2019habitat, andrew_habitat2}.

\begin{figure}[t]
\centering
\includegraphics[width=\linewidth]{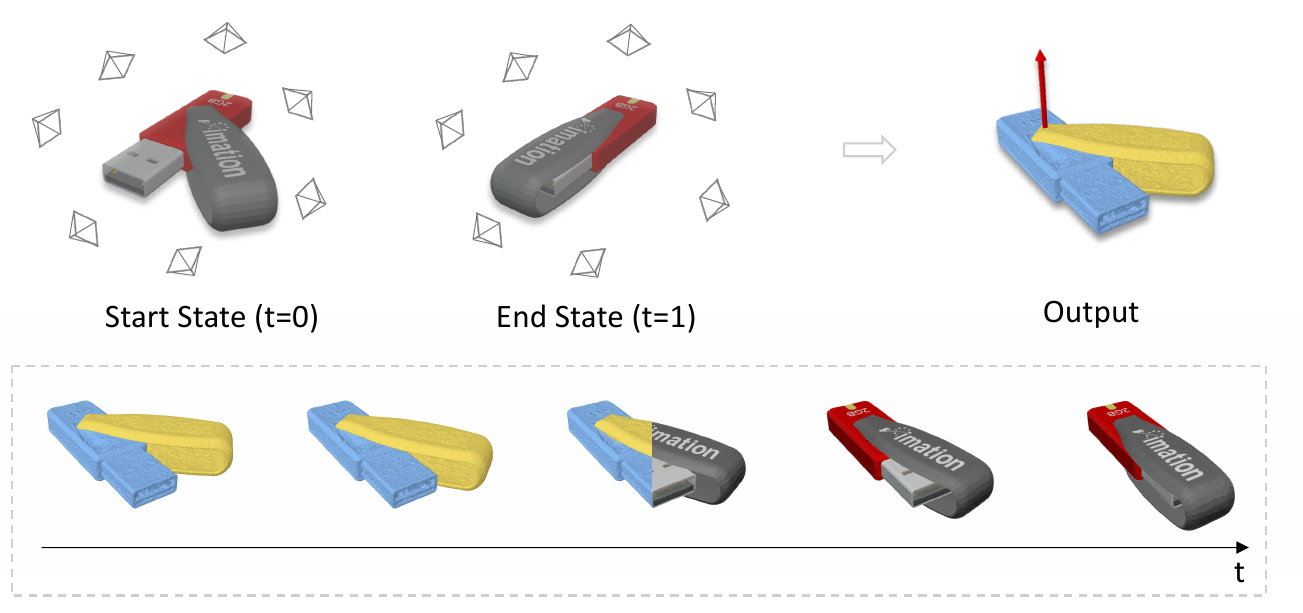}
\caption{We present \paris: a method that takes only multi-view images of an object in two articulation states (top left) and recovers part-level shape and appearance while jointly estimating articulation motion parameters (top right). Our method can generate unseen states and render the object from arbitrary views (bottom).}
\label{fig:task}
\end{figure}

Prior work on articulated object understanding uses supervised learning which requires 3D supervision and articulation annotation~\cite{wang_siga18, yan_rpmnet}.
Unfortunately, such supervisory data is expensive and unavailable at scale.
Another line of prior work assumes a known object category and learns separate models for each category~\cite{li2019category, mu2021asdf, wei_self-supervised_2022, tseng_cla-nerf:_2022}.
This makes generalization to arbitrary unseen objects difficult.
Recently, \citet{jiang_ditto:_2022} proposed Ditto: a category-agnostic approach for motion and part geometry prediction from a pair of 3D point clouds.
However, this approach is limited in generalization to unseen object categories and it does not address detailed appearance reconstruction.

We make the observation that perception of articulated objects involves two subproblems: \emph{reconstruction} and \emph{motion analysis}.
These subproblems are tightly intertwined since knowing the complete geometry of an object makes motion analysis easier, while knowing the motion parameters of an object provides a signal for better reconstruction of the object given observations in different articulation states.
Our insight is that by leveraging the intertwined nature of articulated object perception we can avoid reliance on explicit 3D data and motion parameter supervision.

In this paper, we propose \emph{\paris}: a self-supervised approach for joint reconstruction and motion analysis of articulated objects.
By observing an articulated object in two states (see \Cref{fig:task}), our method reconstructs the shape and appearance of the static and movable parts in two implicit neural fields, while predicting the articulation motion parameters.
The separated neural fields are composited using the estimated motion parameters to set up self-supervisory losses relying only on the input RGB images.

Thus, our approach is category-agnostic and does not require any 3D data or supervisory signals for part segmentation, motion parameters, or object category semantics.

In summary:
\begin{itemize}\denselist
    \item We address joint reconstruction and motion analysis for articulated objects including part-level shape and appearance, and motion estimation given RGB images of the object in only two static states.
    \item We develop \paris: a \emph{category-agnostic}, \emph{self-supervised} and \emph{end-to-end} approach that jointly performs reconstruction and motion analysis without 3D supervision, motion parameter or semantic annotation.
    \item We systematically evaluate our approach on synthetic and real data, and show we significantly outperform prior work and baselines in shape and appearance quality, and motion parameter estimation accuracy.
\end{itemize}

%% file: sections/02_related_work.tex
\section{Related Work}

\textbf{Movable part segmentation and analysis.}
The analysis of part mobility is a well-established challenge for understanding the kinematics of articulated objects~\cite{pmlr-v100-abbatematteo20a,mo_where2act, haresh_video, Gadre_2021_ICCV}. With more 3D data and annotations of articulation being collected for articulated objects, recent works favor tackling this problem from 3D inputs in a data-driven fashion. 
ScrewNet~\cite{osti_10279367} uses a recurrent neural network to predict articulation without part segmentation, from a sequence of depth images.
Assuming part segmentation is given, \citet{hu_learning_2017} estimate the part mobilities by mapping a point cloud to a class of motion sequences via metric learning.

Although many approaches have been proposed to conduct semantic segmentation on 3D shapes~\cite{qi_pointnet,zhao2021point}, the obtained part segmentation does not necessarily conform to mobilities. Considering this obstacle, later works address the mobility part segmentation and analysis collectively. 
Taking a single point cloud as input, Shape2Motion~\cite{wang_siga18} and \citet{li2019category} propose to learn a category-level model to address this coupled task in a supervised way.
These models have only limited generalization to arbitrary unseen objects since a separate trained model is required for each category.
To mitigate dependency on the object category, 
\citet{yan_rpmnet} and \citet{Abdul_LearningTI} design cross-category networks to predict part segmentation and kinematic hierarchy from a point cloud.
\citet{SWMP23} learn part motion parameters from an over-segmented 3D scan in a semi-supervised manner.
\citet{chu2023cart} proposes a category-agnostic method to manipulate the articulated parts to predefined states driven by a user command.


The above works all focus on understanding the articulation and 3D structure of a point cloud. 
Our work reconstructs part-level surface and appearance jointly with motion estimation from only RGB images, which are more readily available compared to 3D or depth inputs.
The closest work is Ditto~\cite{jiang_ditto:_2022} which also produces part-level surface and articulation parameters by observing two states of the object as the input.
Another similar concurrent work is CARTO~\cite{heppert2023carto} which reconstructs object surfaces with motion estimated from stereo images.
The main differences are the following:
1) Ditto takes a pair of 3D point clouds as input, while we use two sets of multi-view RGB images of the object in two states;
2) Ditto and CARTO only produce geometry (no texture or other surface appearance) and CARTO cannot reconstruct parts;
3) The work mentioned above requires 3D supervision and articulation annotation during training, while our approach is self-supervised only with images.

\textbf{Implicit models for articulated objects.}
Neural implicit models are increasingly popular because of their continuous and topology-free representation.
In early works, shape and deformation modeling of articulated objects with implicit functions requires 3D supervision e.g., NASA~\cite{deng2020nasa} for human objects and A-SDF~\cite{mu2021asdf} for general articulated objects.
With the success of differentiable rendering techniques, shape and appearance models can be learned from multi-view RGB images.
This enables reconstruction of static scenes~\cite{mildenhall2020nerf,wang2021neus,Barron_mipnerf360}, 
rigidly moving objects~\cite{yuan2021star},
as well as deformable objects~\cite{2021narf, park2021hypernerf,Guo_2022_NDVG_ACCV,gao_2020tetrahedral}, and dynamically changing scenes~\cite{pumarola2020d_dnerf,tretschk_nonrigid,wu2022d2nerf}.
Following up on A-SDF~\cite{mu2021asdf}, \citet{wei_self-supervised_2022} proposed a category-level shape and appearance representation for general articulated objects.
Conditioning on an articulated latent code, the network can recover the underlying shape and appearance of an unseen object and generate articulated states by interpolating in the latent space.
Similarly, with 2D segmentation maps and annotation of articulation as extra supervision, CLA-NeRF~\cite{tseng_cla-nerf:_2022} can output additional 2D segmentation and estimate part pose via inverse rendering as post-processing. 

What differentiates our work from the above is that we decouple the articulated parts in both shape and appearance without knowing the object category.
Simultaneously, we estimate motion parameters in an end-to-end manner so that we can explicitly manipulate the articulated object to unseen state.
STaR~\cite{yuan2021star} and D$^2$NeRF~\cite{wu2022d2nerf} share the same strategy in decoupling two components by learning separate fields using motion as a cue but they focus on modeling dynamic scenes from an RGB video, whereas we take two sets of RGB images of the object in different (non-dynamic) states.
Practically, our setting is more scalable as observations of common articulated objects in different states emerge naturally without operator intervention (e.g., a folding chair on two different days, once when used and once when put away folded).
This setting introduces more challenges as our input is more sparse and exhibits occlusions both across views (tightly connected parts) and between states (e.g., in \Cref{fig:task} end state a large portion of the static part is occluded).

%% file: sections/03_problem_statement.tex
\begin{figure*}[t]
\begin{center}
\includegraphics[width=\linewidth]{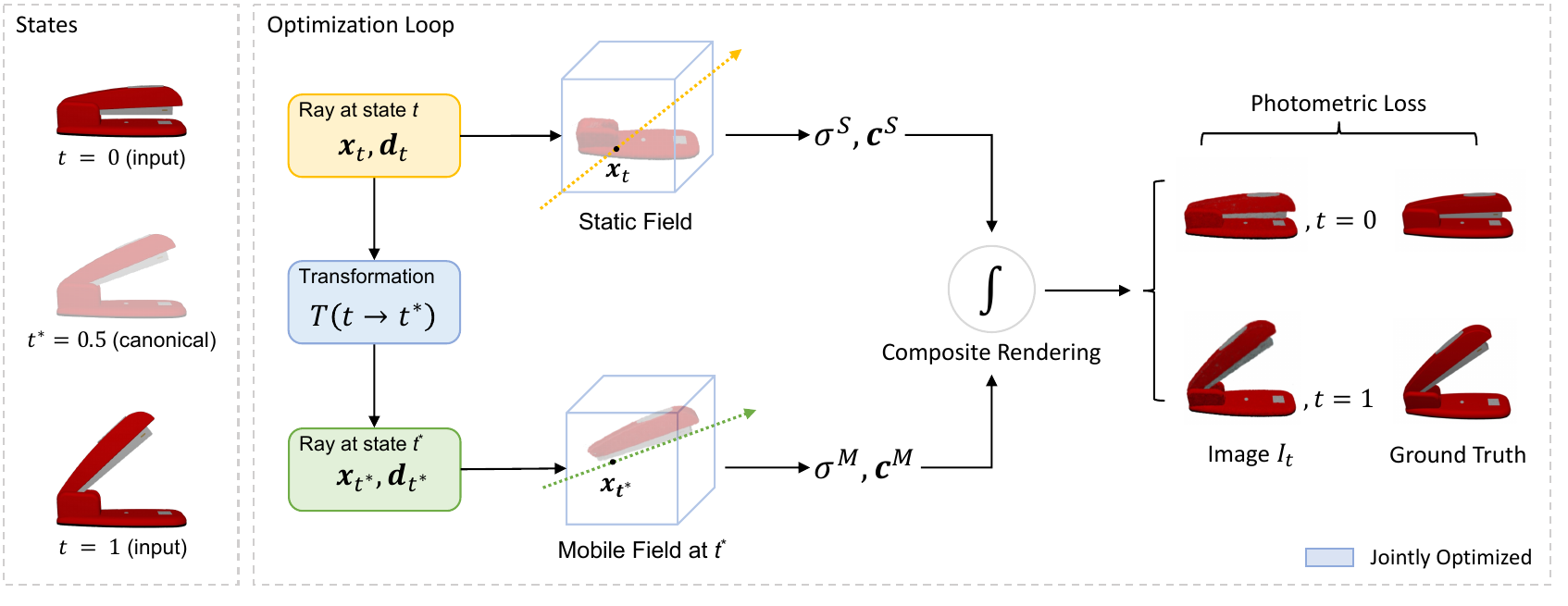}

\end{center}
\vspace{-2mm}
\caption{Method overview.
We learn a part-level reconstruction in two separate fields through composite rendering supervised by images in two given states simultaneously. 
To render an image at input state $t$, we query the static field with the rays at $t$ to obtain the static part, then we transform the ray from $t$ to $t^*$ to query the mobile field which returns the corresponding points of movable part at $t$ from its canonical state. By compositing the colors from two fields, we can supervise the rendering results with the ground truth images from the input state $t$. The network and motion parameters will be jointly optimized for each specific scene. 
}
\label{fig:pipeline}
\end{figure*}

\section{Problem Statement}
\label{sec: prob_state}

Considering an articulated object from an unknown category, our input is composed of two arbitrary articulation states of the object: start state $t=0$ and end state $t=1$. At each state $t$, a set of multi-view RGB images $I^t$ with corresponding camera parameters are given. We assume only one part is moving in this pair of observations, where we call the moving part the \emph{movable} part and the part remaining still the \emph{static} part. Our first goal is to decouple the two parts in terms of both geometry and appearance. With a part-level shape and appearance model in hand, we can articulate the object to unseen states and render the object in new states from arbitrary views easily with articulated motion control. 

Our second goal is followed by articulated motion estimation. 
We assume the movable part exhibits either rotation or translation only.
The motion type is required to estimate the motion parameters (joint axis and joint state).
If it is not given, we first optimize the transformation as a $\mathbf{SE}(3)$ group to classify the motion type as a pre-processing step.
Once the motion type is known we parameterize the joint accordingly.
For a revolute joint, we parameterize it with a pivot point $\textbf{p}\in \mathbb{R}^3$ and a rotation in the form of a unit quaternion $\textbf{q}\in\mathbb{R}^4, \|\textbf{q}\|=1$. For a prismatic joint, we parameterize it with the joint axis as a unit vector $\textbf{a}\in \mathbb{R}^3, \|\textbf{a}\|=1$ and translation distance $d$ along this axis. Now we have a rotation function $f_{\textbf{p}, \textbf{q}}$ and a translation function $f_{\textbf{a}, d}$ using these parameters.
Depending on the motion type given, one of the transformation function $T\in\{f_{\textbf{p}, \textbf{q}}, f_{\textbf{a}, d}\}$ will be plugged into the training pipeline to optimize the motion parameters jointly.

%% file: sections/04_methods.tex
\begin{figure*}[t]
\begin{center}
    \includegraphics[width=\linewidth, trim={6mm 2mm 6mm 2mm}, clip]{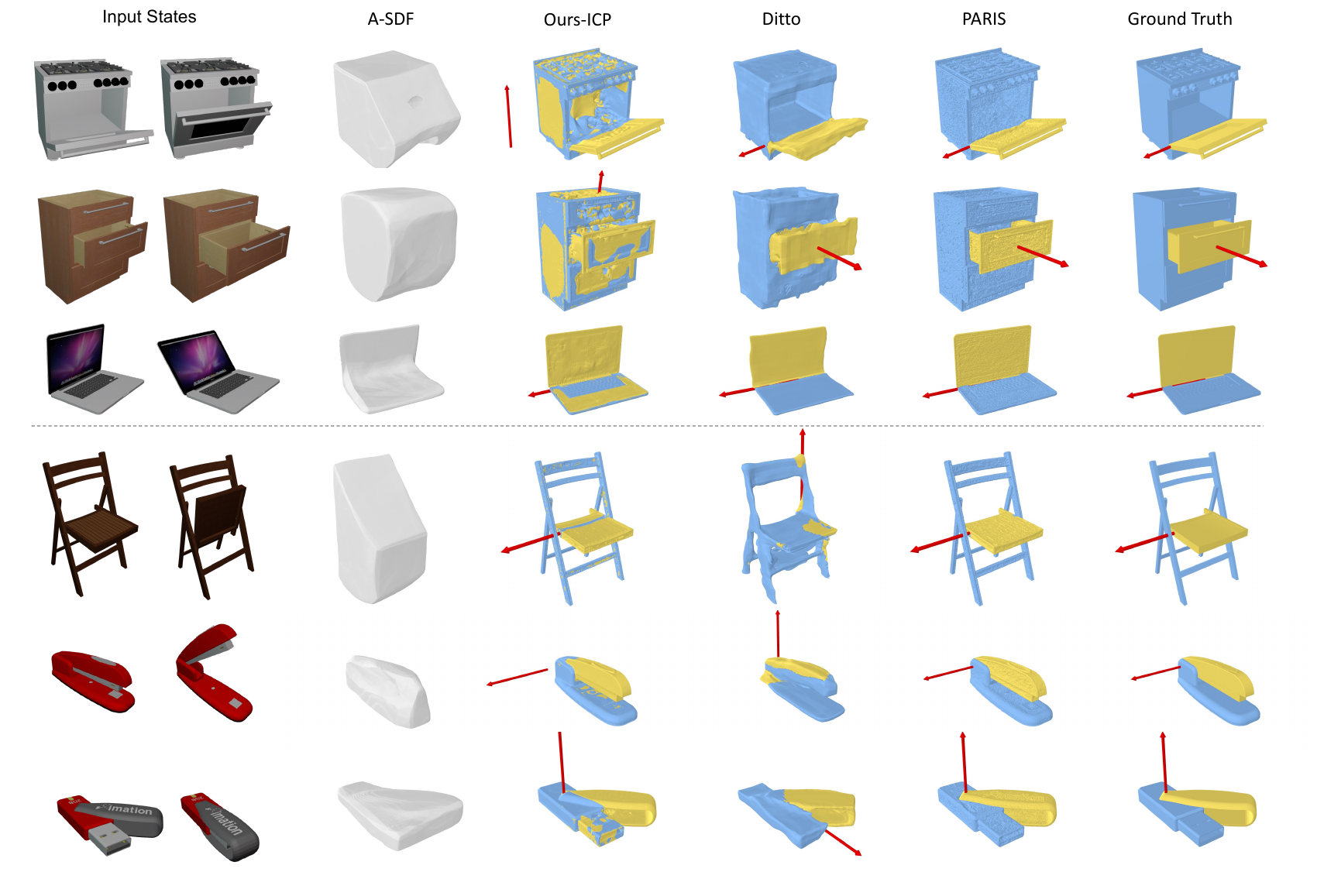}
\end{center}
\caption{Qualitative results for the object and part reconstruction with joint axis visualization. Static parts are colored blue while movable parts are colored yellow. Ditto performs well in both segmentation and axis prediction in the three seen categories (above the dashed line), but the quality drops significantly for unseen categories (three cases below the dashed line). Our methods produce better segmentation with more geometric details while predicting accurate axis across all the categories.
}
\label{fig:quali-geometry}
\end{figure*}

\section{Method}
Our method is an end-to-end framework that learns a part-level representation and estimates articulation jointly from object-level observations in a fully self-supervised fashion.
We separate the static and movable parts by leveraging motion as a cue.
Since the motion accounts for the inconsistency between two states, we optimize the motion parameters by registering the moving parts from the input state $t$ to a canonical state $t^*$. During registration, the component that agrees with the transformation is extracted as the movable part. And the one remaining still is extracted as the static part.
\Cref{fig:pipeline} illustrates the design of our pipeline.


Next, we explain our architecture (\Cref{subsec:composite_fields}), and the supervision losses (\Cref{subsec:loss}) in detail.

\subsection{Composite Neural Radiance Fields}
\label{subsec:composite_fields}

We learn the static and mobile fields compositely during training, and the two fields share the same network architecture which is built upon Instant-NGP~\cite{mueller2022instant}. 
Note that we design each field to model a static scene (one part in one state), meaning neither of the fields is conditioned on the state $t$ as the input. We are not modeling the continuous dynamics for the movable part since we only have multi-views of the object at two discrete states for supervision.
We build the motion relation between two states by learning an explicit transformation function (defined in \Cref{sec: prob_state}) that maps a canonical state to the two input states instead of relying on learned dynamics embedded in the field implicitly, which differentiates our method from other works studying the reconstruction of the dynamic scenes~\cite{wu2022d2nerf,pumarola2020d_dnerf}.
Theoretically, any state $t$ 
can be chosen for the movable part.
To balance the signal backpropagated from the loss on the two input states, we use $t^*=0.5$ as a canonical state to be learned in the field for the movable part.
See the supplement for details about the choice of canonical state.

The static field $\mathcal{F}^S$ represents the static part that remains still in any state, and the mobile field $\mathcal{F}^M$ represents the movable part in the canonical state ($t^*=0.5$).
Formally, they are represented as $\mathcal{F}^S(\textbf{x}_t, \textbf{d}_t)=\sigma^S(\textbf{x}_t), \textbf{c}^S(\textbf{x}_t, \textbf{d}_t)$ and $\mathcal{F}^M(\textbf{x}_{t^*}, \textbf{d}_{t^*})=\sigma^M(\textbf{x}_{t^*}), \textbf{c}^M(\textbf{x}_{t^*}, \textbf{d}_{t^*})$
where $\textbf{x}_t\in\mathbb{R}^3$ is a point sampled along a ray at state $t$ with direction, $\textbf{d}_t\in\mathbb{R}^3$. 
$\sigma(\textbf{x})\in\mathbb{R}$ is the density value of the point $\textbf{x}$, and $\textbf{c}(\textbf{x}, \textbf{d})$ is the RGB color predicted from the point $\textbf{x}$ from a view direction $\textbf{d}$. 
For an input state $t\in\{0, 1\}$, we transform the point from $\textbf{x}_t$ to $\textbf{x}_{t^*}$ and the view direction from $\textbf{d}_t$ to $\textbf{d}_{t^*}$ to query the field $\mathcal{F}^M$, using the transformation function $T$ which is defined in \Cref{sec: prob_state}.
Learning an accurate articulation and segmentation establishes a correspondence from state $t$ to $t^*$.


We adopt a similar training pipeline to the original NeRF~\cite{mildenhall2020nerf} and extend the ray marching and volumetric rendering procedure to compose the two fields. For each training iteration, we independently sample points to produce rendering for each input state $t\in\{0, 1\}$ and supervise the results respectively. During ray marching for each state $t$, we first uniformly sample points $\textbf{x}_t$ along each ray to query the field $\mathcal{F}^S$, then we transform $\textbf{x}_t$ to $\textbf{x}_{t^*}$ to query the field $\mathcal{F}^M$. The summation of the output densities $\sigma_{sum} = \sigma^S(\textbf{x}_t) +\sigma^M(\textbf{x}_{t^*})$ is used to construct the probability density function to guide the stratified sampling in the second round of ray marching. 

\input{tables/surface_v2}

Let $\textbf{x}(h) = \textbf{o}+h\textbf{d}$ be a point along a ray $\textbf{r}=\textbf{o}+h\textbf{d}$ emitted from the center of the projection $\textbf{o}$ with direction $\textbf{d}$. Considering a near and far bound $[h_n, h_f]$, we composite the output from two fields to calculate the color $\hat{C}(\textbf{r})$ for the ray $\textbf{r}$ by integrating the weighted sum of the two colors for each point along the ray:
\begin{equation}
    \hat{C}(\textbf{r}) = \int_{h_n}^{h_f} \left(w^S(h)\cdot\textbf{c}^S(h) + w^M(h)\cdot\textbf{c}^M(h)\right) dh             
\end{equation}
where we simplify our notation as $\textbf{c}(h)\equiv\textbf{c}(\textbf{x}(h), \textbf{d})$.
We define the weights $w^S(h) = T(h)\cdot\alpha^S(h)$ and $w^M(h) = T(h)\cdot\alpha^M(h)$,
where $T(h)$ is the transmittance at the point $\textbf{x}(h)$ accumulated from the two fields with the summation density $\sigma_{sum}(s)$. This is defined as:
\begin{equation}
    T(h) = \exp{\left(-\int_{h_n}^h \sigma_{sum}(s)\cdot\delta_s \hspace{0.5em} ds\right)}
\end{equation}
where $\delta_s=h_{s+1}-h_s$ is the distance between adjacent samples. 
The intuition of this additive composition is that sample points from either field with a high-density value can terminate the ray during rendering.
This strategy is also adopted in STaR~\cite{yuan2021star} and D$^2$NeRF~\cite{wu2022d2nerf}.

\subsection{Supervision Losses}
\label{subsec:loss}

We supervise the learning of motion parameters and neural radiance fields jointly with two shared loss functions. The overall loss function is defined as $\mathcal{L}=\mathcal{L}_\text{rgb}+\lambda_\text{mask}\mathcal{L}_\text{mask}$
where the optimizer for motion parameters only considers $\mathcal{L}_\text{rgb}$ and $\mathcal{L}_\text{mask}$ during training. $\mathcal{L}_\text{rgb}$ defines the photometric loss between the rendering RGB value with ground truth RGB value $C(\textbf{r})$ for a pixel hit by ray $\textbf{r}$:
\begin{equation}
    \mathcal{L}_\text{rgb} = \|\hat{C}(\textbf{r})-C(\textbf{r})\|_2^2.
\end{equation}
The mask loss $\mathcal{L}_\text{mask}$ is defined on the opacity $O(\textbf{r})$ for each pixel with a binary mask $M(\textbf{r})$ of the object:
\begin{equation}
    \mathcal{L}_\text{mask} = \text{BCE}\left(O(\textbf{r}), M(\textbf{r})\right)
\end{equation}
where $\text{BCE}$ is the binary cross-entropy loss and $\lambda_\text{mask}=0.1$ during training. The opacity $O(\textbf{r})$ is computed as:
\begin{equation}
    O(\textbf{r}) = \int_{h_n}^{h_f} w(h)  \hspace{0.5em} dh.
\end{equation}

We empirically find that the mobile field easily accumulates noise in regions overlapping with the static field, especially when the static part is a large volume. The static part hides noise inside its volume during composition and we lack supervision on the points behind the surface with volumetric rendering.
To alleviate this problem, we add a regularization term $\mathcal{L}_\text{prob}$.
The intuition is to encourage the composited color for each ray to be only contributed from one field instead of both.
We define the ratio that the color of a ray $\textbf{r}$ contributed from the mobile field as a fraction of the opacity values:
\begin{equation}
P_M(\textbf{r})=\frac{O^M(\textbf{r})}{O^M(\textbf{r})+O^S(\textbf{r})}.
\end{equation}

Then we define our regularization term $\mathcal{L}_{prob}$ on this ratio for each ray as:
\begin{align}
    \mathcal{L}_\text{prob} &= H(P_M(\textbf{r})), \\
    H(x) &= x\cdot\log(x) + (1-x)\cdot\log(1-x).
\end{align}
By minimizing $\mathcal{L}_\text{prob}$ for all the pixels in each iteration, we force $P_M(\textbf{r})$ to be close to either 0 or 1.
Thus, the density for each position will be encouraged to accumulate in only one field.
We weigh this regularization term with $0.001$ and only apply it to the loss for optimizing parameters for implicit fields as it does not provide a meaningful signal for motion estimation.

%% file: tables/surface_v2.tex
\begin{table*}[t]
\resizebox{\linewidth}{!}{
\begin{tabular}{@{}ccccccccccccc@{}}
\toprule
& & \multicolumn{7}{c}{Unseen Category}  & \multicolumn{3}{c}{Seen Category}   \\ 
\cmidrule(l){3-9}\cmidrule(l){10-12}
Metrics& Methods& Stapler   & USB             & Scissor         & Fridge        & FoldChair     & Washer          & Blade           & Laptop        & Oven          & Storage       & \textbf{Mean} \\ \midrule
\multirow{4}{*}{\begin{tabular}[c]{@{}c@{}}CD-w$\downarrow$\end{tabular}}            
& A-SDF    & 14.19          & 7.14            & 10.61           & 13.71         & 40.85         & 12.50           & 3.31            & 2.11          & 21.37         & 22.57           & 14.84 \\
& Ditto    & 2.38           & 2.09            & 1.70            & \textbf{2.16} & 6.80          & \textbf{7.29}   & 42.04           & 0.31          & \textbf{2.51} & \textbf{3.91}   & 7.19      \\
& Ours-ICP & \textbf{0.91}  & 1.97            & 0.52            & 3.44          & \textbf{0.39} & 13.98           & \textbf{0.42}   & 1.54          & 8.40          & 7.67            & \textbf{3.92}  \\
& \paris   & 0.96	        &\textbf{1.80}	  & \textbf{0.30}	& 2.68	        & 0.42	        & 18.31	          & 0.46	        & \textbf{0.25}	& 6.07	        & 8.12	          & 3.94 \\
\midrule          
\multirow{3}{*}{\begin{tabular}[c]{@{}c@{}}CD-s$\downarrow$\end{tabular}}            
& Ditto    & 41.64          & 2.64           & 39.07          & 3.05           & 33.79         & 10.32           & 46.90          & 0.25           & \textbf{2.52} & \textbf{9.18} & 18.94      \\
& Ours-ICP & 1.09           & \textbf{2.37}  & 0.58           & 3.79           & 2.34          & \textbf{8.26}   & \textbf{0.50}  & 0.42           & 5.95          & 12.59         & \textbf{3.79} \\
& \paris   & \textbf{0.94}	& 2.60	         & \textbf{0.28}  & \textbf{2.88}  & \textbf{0.20} & 19.45 	         & 0.58	          & \textbf{0.15}  & 6.19	       & 11.76         & 4.50  \\

\midrule
\multirow{3}{*}{\begin{tabular}[c]{@{}c@{}}CD-m$\downarrow$\end{tabular}}           
& Ditto    & 31.21          & 15.88         & 20.68         & \textbf{0.99}     & 141.11        & 12.89           & 195.93         & 0.19            & 0.94             & \textbf{2.20}   & 42.20  \\
& Ours-ICP & 8.13           & 2.09          & 18.19         & 129.48            & 38.31         & 74.76           & 20.49          & 67.90           & 43.20            & 156.81          & 55.94    \\
& \paris   & \textbf{0.85}	& \textbf{0.89}	& \textbf{0.23} & 1.13	            & \textbf{0.53} & \textbf{0.27}	  & \textbf{5.13}  & \textbf{0.14}	 & \textbf{0.43}	& 20.67           & \textbf{3.06} \\
\bottomrule
\end{tabular}
}
\vspace{1pt}
\caption{Quantitative results for the surface quality of the reconstructed object and parts. We outperform other methods on the reconstruction of the whole object and movable part on average, and we have comparable quality with \textit{Ours-ICP} on the static part reconstruction.
}
\label{tab:quant-surf}
\end{table*}

%% file: sections/05_experiments.tex
\section{Experiments}


\input{tables/joint_axis_v2}
\input{tables/joint_state_v2.tex}

\subsection{Dataset}

\mypara{Synthetic dataset.}
The synthetic 3D models we use for evaluation are from the PartNet-Mobility dataset~\cite{Xiang_2020_SAPIEN,Mo_2019_CVPR,chang2015shapenet}, a large-scale dataset for articulated objects across 46 categories.
We select instances across 10 categories to conduct our experiments.
For each articulation state, we randomly sample 64-100 views covering the upper hemisphere of the object to simulate capturing in the real world.
Then we render RGB images and acquire camera parameters and object masks using Blender~\cite{blender} to create our training data.

\mypara{Real-world dataset.}
The real data we use for experiments is from the MultiScan dataset~\cite{mao2022multiscan}, scanning real-world indoor scenes with articulated objects in multiple states.
We use the reconstructed mesh of an object in two states as ground truth for evaluation, and the real RGB frames as training data.

\subsection{Baselines}

\mypara{A-SDF.} 
A-SDF~\cite{mu2021asdf} learns a category-level model that reconstructs object meshes given ground truth SDF samples. To compare the generalization ability of the model across categories, we train the model on the 10 testing examples and retrieve them during testing. We follow the method that DeepSDF~\cite{Park_2019_CVPR} proposed to generate SDF samples for each testing case. Note that A-SDF cannot predict part-level geometry or estimate explicit motion parameters, we only measure the quality of surface reconstruction on the whole object for comparison.

\mypara{Ditto.}
Ditto~\cite{jiang_ditto:_2022} learns a category-agnostic model that reconstructs part meshes and estimates motion parameters (motion type, joint axis, joint state) from a pair of 3D point clouds. We leverage their released pre-trained model which is trained on Shape2Motion~\cite{wang_siga18} dataset across 4 categories to do the comparison. And we test instances across 10 categories (3 seen and 7 unseen by Ditto, marked in \Cref{tab:quant-surf}) to compare the generalization ability regarding the part-level surface reconstruction and motion estimation. We sample point clouds on the surface of the ground truth meshes of the object in two given articulation states to feed as their input for both synthetic and real data experiments.

\begin{figure}[t]
\centering
\includegraphics[width=0.95\linewidth, trim={1mm 1mm 2mm 7mm}, clip]{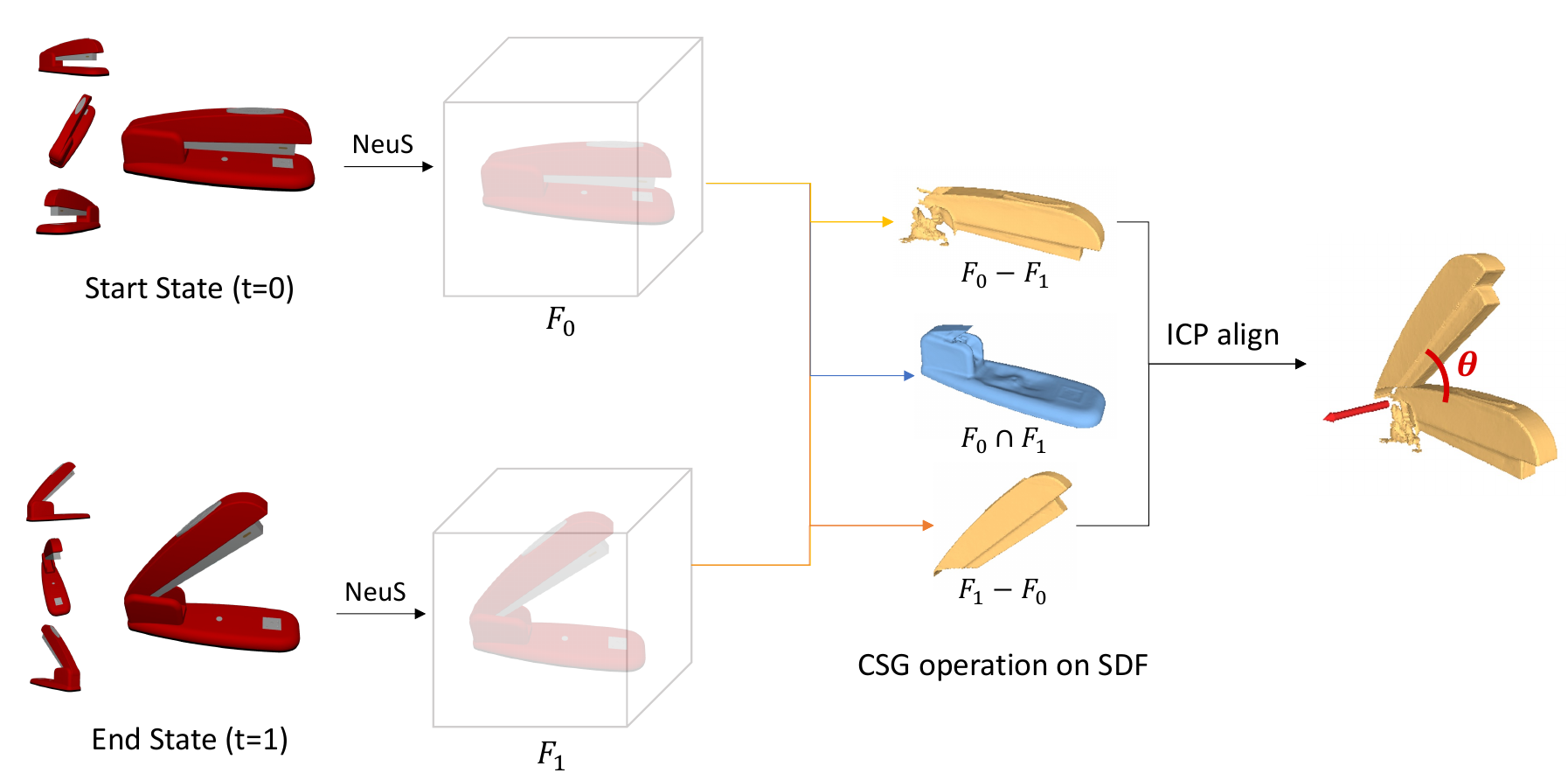}
\caption{Pipeline of our baseline method \textit{Ours-ICP}. We learn separate neural SDFs $F_0, F_1$ for the object in state $t=\{0, 1\}$. To segment static and movable parts, we apply CSG operations between two SDFs. Then we leverage global registration and the ICP algorithm to obtain the transformation on the movable parts from the start to end state.}
\label{fig:baseline}
\vspace{-1mm}
\end{figure}

\mypara{Ours-ICP.}
None of the baseline methods mentioned above can produce the appearance of the object or even render the object in arbitrary views since they did not consume RGB information as their input. Here, we implemented a naive baseline approach as illustrated in \Cref{fig:baseline}. We learn two separate neural implicit surfaces $F_0, F_1$ from the two given sets of multi-view images, which are backboned with NeuS~\cite{wang2021neus} and Instant-NGP~\cite{mueller2022instant}. Then we compare the quality of the novel view synthesis at two given articulated states with this baseline model. Additionally, we apply Constructive Solid Geometry (CSG) algorithm between the two neural SDFs and extract the part-level geometry with Marching-Cube~\cite{loren_marchcube}. Specifically, the intersection region of the two fields can be extracted as the static part. By subtracting one field from the other, we can obtain the movable part in two states accordingly. We can further compute the transformation between the movable parts in two states by using global registration~\cite{rusu_fpfh2009,fischler_ransac,sungjoon_choi_robust_2015} and generalized ICP registration~\cite{Segal2009GeneralizedICP} algorithms.

\subsection{Evaluation Metrics}

\mypara{Part-level geometry.}
We use the Chamfer-L1 distance (CD) as the metric to evaluate the quality of the reconstructed meshes. We measure the CD on the whole reconstructed surface (CD-w) to compare it with all the baselines. We also measure the CD on the predicted static part (CD-s) and movable part (CD-m) separately to compare them with all the baselines except for A-SDF since it cannot predict part segmentation without ground truth SDFs for parts. 
To be more specific, we sample 10,000 points on each surface to compute the distance from both the prediction to the ground truth and the other way around, then we average the two distances to report as the final metrics. The CD values shown are multiplied by 1,000 as in A-SDF and Ditto.

\mypara{Motion estimation.}
To evaluate the joint axis, we measure the angular error (Ang Err) for both types of joints. This metric computes the orientation difference between the predicted axis direction and the ground truth, ranging from 0 to 90 degrees.  We also measure the position error (Pos Err) for revolute joints. This metric computes the minimum distance between the rotation axis and the ground truth which takes the position of the pivot point into account. The value is multiplied by 10 shown in \Cref{tab:quant-axis}. To evaluate the joint state, we measure the geodesic distance (in degree) for revolute joints and translation error for prismatic joints.

\mypara{Novel view synthesis.}
To evaluate the quality of the appearance model, we measure the Peak Signal-to-noise Ratio (PSNR) and the Structural Similarity Index (SSIM)~\cite{zhou_ssim} for rendered results from novel views. For each instance, we render 50 novel views of the object in each state and average the values.
We achieve comparable rendering quality with the ICP baseline.
See the supplement for details.

\begin{figure}[t]
\centering
\includegraphics[width=\linewidth]{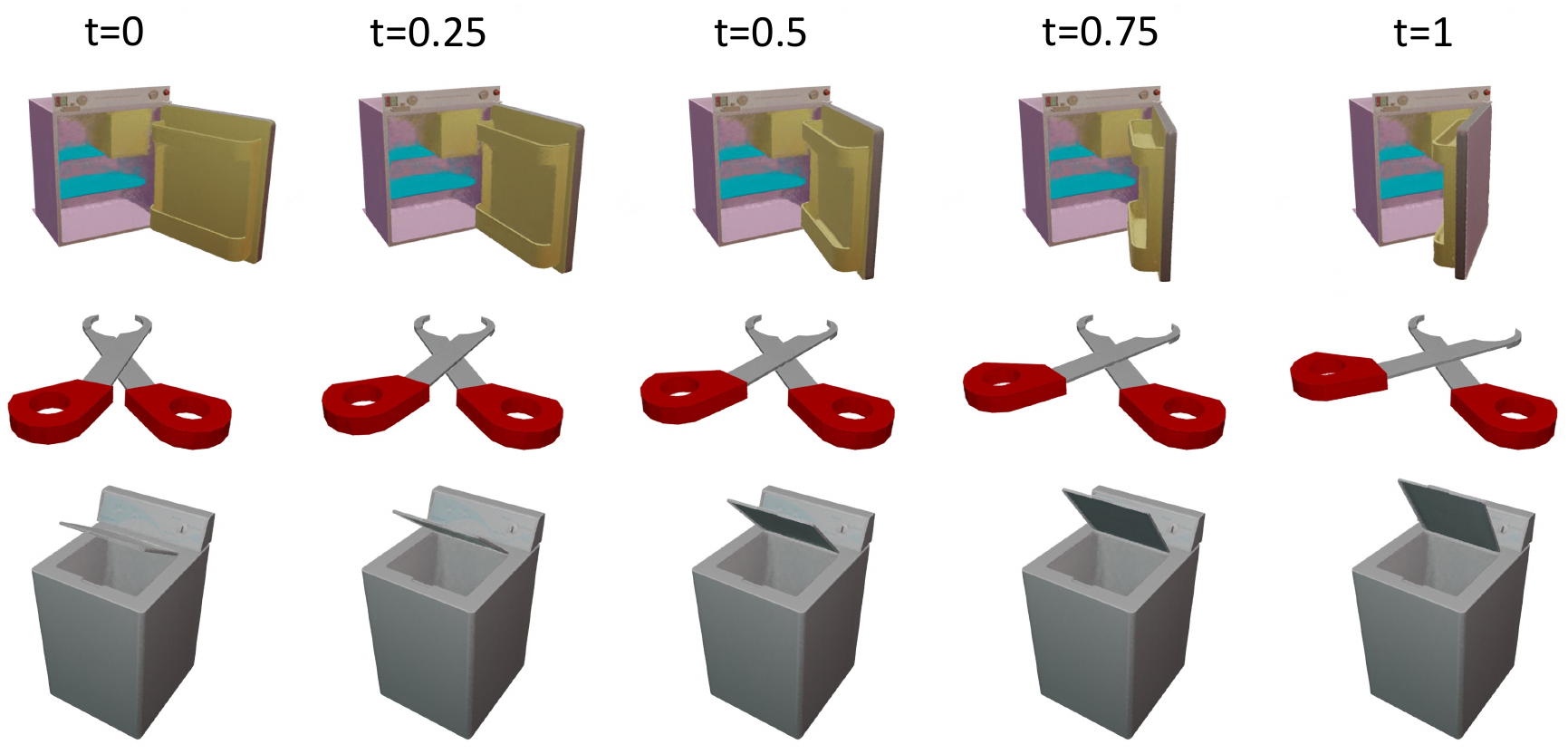}
\caption{Rendering of unseen states from an arbitrary novel view by interpolating $t$ as input. Our high-quality results demonstrate that our motion prediction and decoupling of the part appearance are accurate to effectively generate arbitrary unseen states.
}
\label{fig:quali-states}
\end{figure}

\subsection{Experiment Results}

\mypara{Reconstruction and part segmentation.}
The quantitative results are in \Cref{tab:quant-surf} and a qualitative comparison is in \Cref{fig:quali-geometry}. Note that our method reconstructs the canonical state of the movable part as the output, whereas Ditto~\cite{jiang_ditto:_2022} outputs the start state.
Thus, we transform our predicted movable part from the canonical state to the start state using the predicted motion for these results. 

For the whole object reconstruction, we observe in \Cref{tab:quant-surf} that Ditto outperforms for \emph{Fridge}, \emph{Washer}, \emph{Oven}, and \emph{Storage}. Since Ditto takes ground truth 3D point clouds as input, they have more information about the inner space of these objects than we do with limited image views. So it is understandable that they perform better when a deep container is involved. For part segmentation, we outperform 7 out of 10 cases for both the static and movable parts. We have a big improvement over the movable part on average and outperform the baselines on average. From \Cref{tab:quant-surf}, we observe that Ditto performs well in the three seen categories (above the dashed line), but the quality drops significantly for unseen categories (below the dashed line). Also, we produce more geometric details and better segmentation, which indicates we have a better generalization ability.

\begin{figure*}[t]
\includegraphics[width=\linewidth,trim={0.5cm 1.5cm 0 0.7cm},clip]{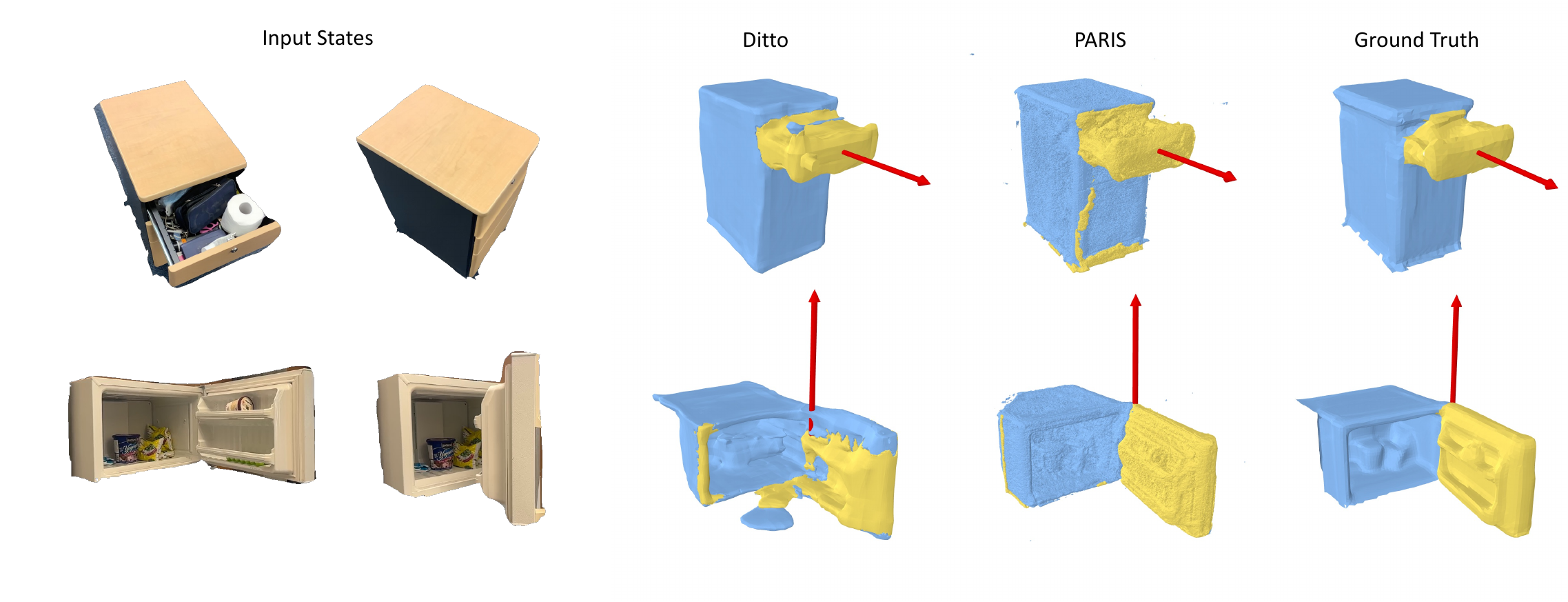}

\caption{Qualitative results for two real-world objects. 
Ditto and our method can both estimate accurate motion parameters.
Ditto produces cleaner segmentation in the first case, but a less accurate geometry in the second case.
}
\label{fig:quali-real}
\end{figure*}

\mypara{Motion estimation.}
We show the quantitative results of the joint axis and joint state estimation in \Cref{tab:quant-axis} and \Cref{tab:quant-state}. The predicted axis is also visualized in \Cref{fig:quali-geometry}. 
We observe that our motion estimation is significantly more accurate than other baselines for both revolute and prismatic joints. Since Ditto predicts a wrong motion type for two of the testing examples, we report a failure (F) in \Cref{tab:quant-state} for joint state evaluation and denote $*$ beside the numbers in \Cref{tab:quant-axis} for joint axis evaluation.

\mypara{Novel view synthesis and articulation generation.}
We show the qualitative results of generating the intermediate articulation using the motion we predict from an arbitrary view in \Cref{fig:quali-states}. Our high-quality rendering results for the intermediate states demonstrate that our decoupling of the part appearance model is accurate, and we can effectively generate arbitrary unseen states and render from arbitrary views with our part-level implicit representation and predicted motion parameters.

\begin{figure}[t]
\centering
\includegraphics[width=\linewidth]{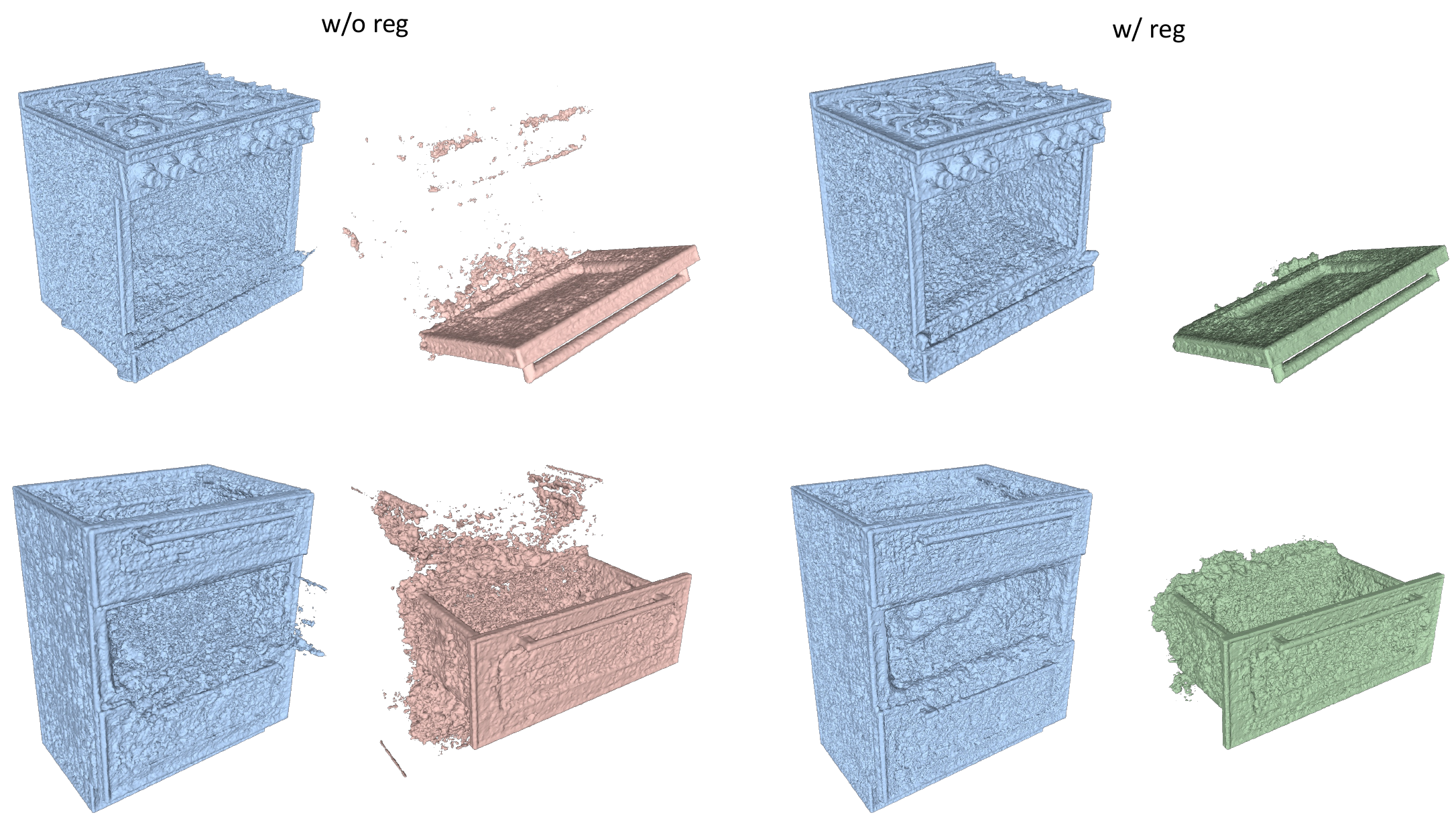}
\caption{Ablation of our regularization term. The left two columns show part reconstruction without this term, and the right two columns show the result with the term added. Note the reduction in movable part noise.
}
\label{fig:quali-ablation}
\end{figure}

\input{tables/ab_reg_v2.tex}

\mypara{Ablation studies.}
We conduct an ablation for the effectiveness of the regularization term in improving reconstruction quality, motion estimation, and appearance. 
The results are in \Cref{tab:ablation}.
We observe that the regularization can indeed improve the surface reconstruction for both part and object on average, especially for the movable part where we do not have sufficient views to observe the whole part (e.g., \textit{Washer, Oven}). We also show qualitative results in \Cref{fig:quali-ablation} to have a closer look at the changes in the movable part for the two most effective cases.

We have an ablation for the choice of canonical state reconstructed in the mobile field. We choose to reconstruct a ``virtual'' state at $t^*=0.5$ in our pipeline.
This better fits our problem statement as it allows us to utilize supervision signals from both given states to predict the motion and segmentation.
In~\Cref{tab:ablation:canonical}, we show the impact on performance of using given state $t^*=0$ (more similar to STaR\cite{yuan2021star}) as the canonical state instead.
We note a performance reduction across all aspects except for appearance for which the results are comparable, with a particularly large degradation in movable part geometric accuracy.

We also carry out an ablation over the number of input views by downsampling the views with farthest point sampling.
In~\Cref{tab:ablation:views}, we reproduce 64 view results from \Cref{tab:quant-surf,tab:quant-axis,tab:quant-state} and report results for 32 to 4 views.
We observe a decline in performance with decreasing views, with a significant drop below 16 views.
This is expected given the challenge of learning NeRFs in the few input regime, and constitutes an interesting direction for future work.

\input{tables/real_v2}
\begin{figure*}[t]
\includegraphics[width=\linewidth]{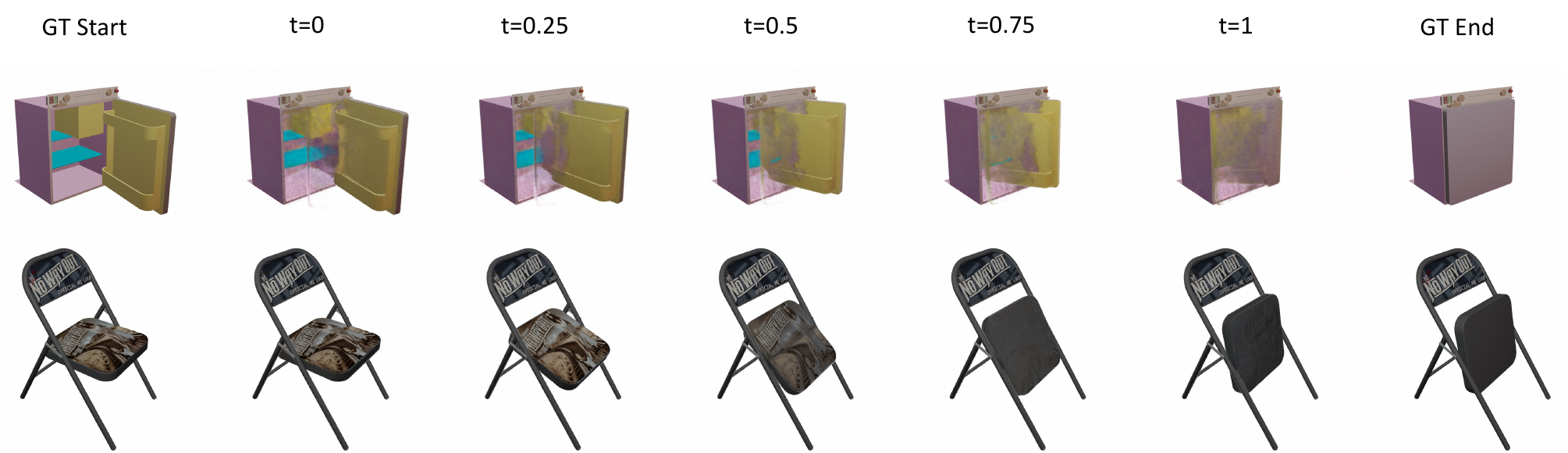}
\caption{Failure cases of articulation generation with incorrectly predicted motion. This figure illustrates how our predicted motion manipulates the movable part to satisfy the supervision in the wrong way. The door of the fridge is wrongly rotating around an axis away from the body, and the seat of the chair is rotating around the correct joint axis but in an opposite direction.}
\label{fig:fail_states}
\end{figure*}

\mypara{Real-world examples.}
Comparison with Ditto for two real cases is shown in~\Cref{fig:quali-real} and~\Cref{tab:real}. Since Ditto and our method both assume that the geometries of the object in two states are pre-aligned, we both suffer from the error in alignment and reconstruction on the given mesh. Ditto and our method can both estimate accurate motion parameters. Ditto produces cleaner segmentation in the first case, but a less accurate geometry in the second case.

\subsection{Limitations}
To obtain an accurate decoupling of the parts and motion parameters, our method relies on a sufficient number of multi-view observations of the object for both static and movable parts. 
Our inputs exhibit occlusions both across views (tightly connected parts) and between states (e.g. in~\Cref{fig:task} end state, a large portion of the static part is occluded). This increases the level of difficulty in finding 3D correspondence from RGB images only.
We observe that our method tends to fail to estimate the correct motion parameters for revolute joints when the movable part is 1) severely occluded across views (e.g., the door of the cabinet is fully closed); or 2) highly geometrically symmetric. 
We illustrate these failure modes in~\Cref{fig:fail_states} by rendering objects from a novel view in sequential states. 
This figure illustrates how our predicted motion manipulates the movable part to satisfy the supervision at two given states in the wrong way. 
Specifically, the door of the fridge is wrongly rotating around an axis away from the body, and the seat of the chair is rotating around the correct joint axis but in an opposite direction.
We also observe that the optimization on the very thin parts is inclined to be unstable which can be alleviated by increasing the number of viewpoints.


\input{tables/ab_canonical}
\input{tables/ab_n_views}

%% file: tables/joint_axis_v2.tex
\begin{table*}[t]
    \begin{center}
    \resizebox{\textwidth}{!}{
        \begin{tabular}{@{}ccccccccccccc@{}}
\toprule
& & \multicolumn{8}{c}{Revolute}  & \multicolumn{2}{c}{Prismatic}   \\ 
\cmidrule(l){3-10}\cmidrule(l){11-12}
Metrics & Methods & Stapler     & USB               & Scissor           & Fridge            & FoldChair         & Washer            & Oven          & Laptop        & Blade          & Storage          & \textbf{Mean}  \\ 
\midrule
\multirow{3}{*}{Ang Err$\downarrow$} 
& Ditto    & 89.86              & 89.77             & 4.498             & 89.30*            & 89.35             & 89.51             & 0.955         & 3.124          & 6.319         & 79.54*              & 54.22    \\
& Ours-ICP & 0.179              & 1.142             & 0.528             & 5.192             & 0.256             & 89.20             & 74.94         & 0.488          & 80.71         & 22.33               & 27.50    \\
& \paris   & \textbf{0.069}	    & \textbf{0.065}	& \textbf{0.019}	& \textbf{0.001}	& \textbf{0.020}    & \textbf{0.082}	& \textbf{0.028}& \textbf{0.034} & \textbf{0.001}& \textbf{0.369}      & \textbf{0.069}    \\
\midrule
\multirow{3}{*}{Pos Err$\downarrow$} 
& Ditto    & 0.201             & 5.409              & 5.698             & 1.021*            & 3.768              & 0.661             & 0.129         & 0.014          & -     & -    & 2.113    \\
& Ours-ICP & 0.073             & 0.183              & \textbf{0.000}    & 2.965             & \textbf{0.000}     & 2.427             & 9.035         & 0.047          & -     & -    & 1.841    \\
& \paris   & \textbf{0.006}	   & \textbf{0.000}	    & \textbf{0.000}	& \textbf{0.002}	& 0.004	             & \textbf{0.02}	 & \textbf{0.003}& \textbf{0.001} & -     & -    & \textbf{0.005}    \\
\bottomrule
\end{tabular}}
    \end{center}
    \caption{Quantitative results for evaluating the error in joint axis prediction. We perform the best across the 10 examples. Ditto achieves four comparable estimations, four wrong ones, and two with the wrong motion type (marked with $*$). Ours-ICP performs in between.
    }
    \label{tab:quant-axis}
\end{table*} 

%% file: tables/joint_state_v2.tex
\begin{table*}[t]
\begin{center}
\resizebox{\textwidth}{!}{
\begin{tabular}{@{}ccccccccccccc@{}}
\toprule
& \multicolumn{8}{c}{Revolute} & & \multicolumn{2}{c}{Prismatic}   \\ 
\cmidrule(l){2-9}\cmidrule(l){11-12}
Methods & Stapler           & USB           & Scissor       & Fridge         & FoldChair        & Washer           & Oven           & Laptop        & \textbf{Mean}    & Blade          & Storage       & \textbf{Mean}  \\ 
\midrule
Ditto    & 56.61            & 80.60         & 19.28         & F              & 99.36            & 55.72            & 2.094          & 5.181         & 50.89            & F              & 0.086          & 0.086     \\
Ours-ICP & 0.423            & 1.704         & 0.364         & 11.39          & 0.465            & 175.1            & 68.48          & 0.668         & 32.32            & 0.262          & 0.311          & 0.287 \\
\paris   & \textbf{0.000}   & \textbf{0.028}& \textbf{0.000}& \textbf{0.001} & \textbf{0.000}   & \textbf{0.079}   & \textbf{0.000} & \textbf{0.028}& \textbf{0.017}   & \textbf{0.064} & \textbf{0.000}& \textbf{0.032}  \\
\bottomrule
\end{tabular}}
\end{center}
\caption{Estimation for the joint state - we measure the geodesic distance $\downarrow$ (degree) for revolute joints and translation error $\downarrow$ for prismatic joints. We can predict significantly accurate rotation angles for revolute joints, and also performs better for prismatic joints. The cases with the wrong motion type predicted by Ditto are denoted as F.
}
\label{tab:quant-state}
\end{table*}

%% file: tables/ab_reg_v2.tex
\begin{table}[t]
\centering
\resizebox{\linewidth}{!}{
\begin{tabular}{@{}llllllllll@{}}
\toprule
& \multicolumn{3}{c}{Geometry} & \multicolumn{4}{c}{Motion} & \multicolumn{2}{c}{Appearance}   \\ 
\cmidrule(l){2-4} \cmidrule(l){5-8} \cmidrule(l){9-10}
Method               
& CD-w$\downarrow$       & CD-s$\downarrow$  & CD-m$\downarrow$   & Ang$\downarrow$  & Pos$\downarrow$   & Geo$\downarrow$  & Trans$\downarrow$ & PSNR$\uparrow$  & SSIM$\uparrow$ \\ \midrule
w/o reg & 4.78         & 4.97              & 23.32              & 0.087            & 0.006             & 0.029            & 0.054           & 36.64   & 0.989 \\
w/ reg & \textbf{3.93} & \textbf{4.50}     & \textbf{5.38}      & \textbf{0.069}   & \textbf{0.005}    & \textbf{0.017}   & \textbf{0.032}  & \textbf{37.73} & \textbf{0.992} \\  \bottomrule
\end{tabular}}
\vspace{1pt}
\caption{Ablation for the regularization term. Geometry, motion, and appearance all improve when using regularization.
}
\label{tab:ablation}
\end{table}

%% file: tables/real_v2.tex
\begin{table}[t]
\centering
\resizebox{\linewidth}{!}{
\begin{tabular}{@{}lllllllll@{}}
\toprule
& & \multicolumn{3}{c}{Geometry} & \multicolumn{4}{c}{Motion}   \\ 
\cmidrule(l){3-5} \cmidrule(l){6-9}
Example & Method               
& CD-w$\downarrow$      & CD-s$\downarrow$  & CD-m$\downarrow$   & Ang$\downarrow$  & Pos$\downarrow$   & Geo$\downarrow$  & Trans$\downarrow$ \\ \midrule
\multirow{2}{*}{Fridge} 
& Ditto & \textbf{6.50} & 47.01             & \textbf{50.60}    & \textbf{1.71}         & 1.84                  & 8.43                  & -\\
& PARIS & 8.20          & \textbf{10.22}    & 67.54             & 1.91                  & \textbf{0.53}         & \textbf{0.77}         & -\\ \midrule
\multirow{2}{*}{Storage}    
& Ditto & \textbf{14.08}& \textbf{16.09}   & \textbf{20.35}    & 5.88                  & -                     & -                     & 0.38 \\
& PARIS & 18.98         & 20.92             & 101.20            & \textbf{3.88}         & -                     & -                      & \textbf{0.31} \\ \bottomrule
\end{tabular}}
\vspace{1pt}
\caption{Quantitative results for real-world objects. We produce comparable quality surfaces for the static part and object as a whole compared to Ditto, but less accurate surfaces for movable parts. Our motion estimation is better.
}
\label{tab:real}
\end{table}

%% file: tables/ab_canonical.tex
\begin{table}[t]
\centering
\resizebox{\linewidth}{!}{
\begin{tabular}{@{}lrrrrrrrrr@{}}
\toprule
& \multicolumn{3}{c}{Geometry} & \multicolumn{4}{c}{Motion} & \multicolumn{2}{c}{Appearance}   \\ 
\cmidrule(l){2-4} \cmidrule(l){5-8} \cmidrule(l){9-10} 
canonical & CD-w$\downarrow$ & CD-s$\downarrow$ & CD-m$\downarrow$ & Ang$\downarrow$ & Pos$\downarrow$ & Geo$\downarrow$ & Trans$\downarrow$ & PSNR$\uparrow$ & SSIM$\uparrow$ \\ \midrule
$t^*=0.5$ & \textbf{3.94}	&\textbf{4.50}	&\textbf{3.06}	&\textbf{0.069}	& \textbf{0.005}	&\textbf{0.017}	&\textbf{0.032}	&\textbf{37.73}	&\textbf{0.992} \\
$t^*=0$ &13.25	&5.64	&74.34	&0.070	& \textbf{0.005}	&0.050	&0.088	&37.62	&\textbf{0.992}\\

\bottomrule
\end{tabular}
}
\vspace{1pt}
\caption{Ablation of canonical state selection. We note a performance reduction across all aspects, with a particularly large degradation in movable part geometric accuracy.}
\label{tab:ablation:canonical}
\end{table}

%% file: tables/ab_n_views.tex
\begin{table}[t]
\centering
\resizebox{\linewidth}{!}{
\begin{tabular}{@{}rrrrrrrrrr@{}}
\toprule
& \multicolumn{3}{c}{Geometry} & \multicolumn{4}{c}{Motion} & \multicolumn{2}{c}{Appearance}   \\ 
\cmidrule(l){2-4} \cmidrule(l){5-8} \cmidrule(l){9-10} 
\# views & CD-w$\downarrow$ & CD-s$\downarrow$ & CD-m$\downarrow$ & Ang$\downarrow$ & Pos$\downarrow$ & Geo$\downarrow$ & Trans$\downarrow$ & PSNR$\uparrow$ & SSIM$\uparrow$ \\ \midrule
64 & \textbf{3.94}	&\textbf{4.50}	&\textbf{3.06}	&\textbf{0.069}	&\textbf{0.005}	&\textbf{0.017}	&\textbf{0.032}	&\textbf{37.73}	&\textbf{0.992} \\
32 & 4.053	& 4.841	 & 19.482	& 0.181	 & 0.041	& 11.346	&0.033	& 34.36	& 0.986\\
16 & 6.677	& 6.479	 & 35.459	& 4.542	 & 0.638	& 14.447	& 0.039	& 30.24	& 0.975 \\
8 & 54.784	& 64.040	 & 196.86	& 36.990 & 2.093	& 29.432	& 0.116	& 19.29	& 0.918 \\
4 & 170.116	& 220.030 & 286.600	& 37.721 & 3.212	& 29.797	& 0.150	& 16.83 & 0.886 \\

\bottomrule
\end{tabular}
}
\vspace{1pt}
\caption{Ablation of input view number. We observe a decline in performance with decreasing views, with a significant drop below 16 views.}
\label{tab:ablation:views}
\end{table}

%% file: sections/06_conclusion.tex
\section{Conclusion}

We addressed the task of joint part-level shape and appearance reconstruction and motion parameter estimation for articulated objects.
Our work is the first to tackle this task from multi-view RGB images observing the object in two arbitrary states.
We evaluated our method on synthetic and real data to systematically study the challenges in this task.
Our experiments show that we can recover shape, appearance, and motion parameters better than prior work and baselines.
However, the task remains challenging especially for cases with severe occlusion. 
Extending to objects with multiple parts is another challenge for future work.
Moreover, we assumed the two given object states are aligned in world coordinates, an assumption shared with prior work.
We hope our work inspires more research into articulated object reconstruction without 3D supervision.